\documentclass[a4paper,conference]{IEEEtran}

\ifCLASSINFOpdf
\else
\fi

\IEEEoverridecommandlockouts

\usepackage{bm}
\usepackage[normalem]{ulem}
\usepackage{graphicx}
\usepackage{multirow}
\usepackage{subcaption}
\useunder{\uline}{\ul}{}
\begin{document}
%
\title{Self-Attention Neural Bag-of-Features}

\author{\IEEEauthorblockN{Kateryna Chumachenko$^1$, Alexandros Iosifidis$^2$ and Moncef Gabbouj$^1$}\thanks{This project has received funding from the European
Union’s Horizon 2020 research and innovation programme
under grant agreement No 871449 (OpenDR).}
\IEEEauthorblockA{$^1$Department of Computing Sciences, Tampere University, Tampere, Finland\\
$^2$Department of Electrical and Computer Engineering, Aarhus University, Aarhus, Denmark\\
Emails: \{kateryna.chumachenko,moncef.gabbouj\}@tuni.fi, ai@ece.au.dk}
}


%



\title{Self-attention fusion for audiovisual emotion recognition with incomplete data}
\maketitle

\begin{abstract}
In this paper, we consider the problem of multimodal data analysis with a use case of audiovisual emotion recognition. We propose an architecture capable of learning from raw data and describe three variants of it with distinct modality fusion mechanisms. While most of the previous works consider the ideal scenario of presence of both modalities at all times during inference, we evaluate the robustness of the model in the unconstrained settings where one modality is absent or noisy, and propose a method to mitigate these limitations in a form of modality dropout. Most importantly, we find that following this approach not only improves performance drastically under the absence/noisy representations of one modality, but also improves the performance in a standard ideal setting, outperforming the competing methods.

\end{abstract}
\section{INTRODUCTION}

Recognition of human emotional states is an important task within the field of machine learning,  enabling better understanding of social signals in the wide range of applications ranging from robotics to human-computer interaction \cite{liu2017facial, chen2019automatic}. Multiple approaches and emotion models have been proposed to date, ranging from the task of recognizing discrete emotional states, such as `happy', `angry', or `sad', to the estimation of emotional attributes, such as arousal and valence on a continuous scale \cite{dzedzickis2020human, mollahosseini2017affectnet}. The task has been approached from multiple angles with different data types used as input, including text \cite{soleymani2017survey}, speech \cite{khalil2019speech}, and images \cite{efficientface}. 

With the abundance of available data, a wide range of methods aiming to fully take advantage of this data are emerging, giving momentum to the development of multi-modal methods \cite{mmtm, chumachenko2021speed, wang2020deep}. Multi-modal methods are a class of methods that operate jointly on multiple data types. These include, among others, video data that consists of audio and visual modalities \cite{xiao2020audiovisual}, joint RGB and depth images \cite{xiao2020multimodal}, and RGB and skeleton data \cite{ignore}. Methods operating on such multi-modal representations range from simple decision-level fusion approaches to more advanced joint feature learning approaches. Although fusion of intermediate features can potentially yield better performance due to joint learning of representations of multiple modalities, late or early fusion remain a popular choice in modern architectures due to their simplicity and versatility \cite{mmtm}. On the other hand, early fusion is not suitable for fusion of drastically different data types, while late fusion primarily only considers features learnt in each modality independently.

Most multi-modal methods developed to date assume full presence of all the adopted modalities at all times during inference, and only evaluate the performance of the models in such a setting. Nevertheless, in real-world applications it is often probable for one of the modalities to be missing or having poor quality at certain times, hence robustness of the model to such scenarios is an essential factor in building multi-modal systems operating on real-world data. 

In the task of multi-modal emotion recognition, especially in the recent transformer-based architectures, the trend has been largely in utilization of pre-extracted features that are subsequently fused with a learnt model, rather than creating end-to-end trainable models \cite{mult, interspeech, icasspav}. This limits the applicability of such methods in real-world scenarios, as necessary feature extraction is often challenging in unconstrained settings and introduces another point of uncertainty to the overall processing pipeline. This is especially the case for the methods adopting language information \cite{mult, interspeech}, as text transcriptions of audio signals are rarely available in practical applications and require separate estimation. We therefore primarily target the task of audiovisual emotion recognition that does not require separate feature learning.

In our work, we aim to address these limitations of existing multi-modal emotion recognition methods by building an end-to-end model that does not require prior feature learning, and performing fusion at intermediate level, while being robust to incomplete or noisy data samples. Our contributions can be summarized as follows: 
\begin{itemize}
    \item We propose a new architecture for audiovisual emotion recognition from facial videos and speech which does not rely on separately learnt features and learns end-to-end from raw videos;
    
    \item We employ several modality fusion approaches and propose an attention-based intermediate feature fusion approach that softly attends to modality-independent features. To the best of our knowledge, such approach has not been proposed before;
    
    \item We propose a new training scheme based on modality dropout mechanisms aimed to improve the robustness of the model under incomplete or noisy data of one modality. We additionally find that the proposed approach yields better performance also in the standard case under presence of both modalities. 
\end{itemize}

\section{Related Work}

Emotion recognition has received a significant amount of attention by the machine learning community, and a number of methods aiming to solve this task have been proposed to date. These methods operate on various data types, such as images \cite{efficientface}, speech \cite{khalil2019speech}, text \cite{soleymani2017survey}, or biosignals \cite{shu2018review}. At the same time, methods combining these modalities have been proposed as well \cite{mult, icasspav, interspeech} employing different multi-modal fusion techniques.

Within the field of multimodal machine learning, generally, three classes of multi-modal fusion approaches are identified: early fusion, where the input data of multiple modalities are simply combined via concatenation, addition, or any other operation and further processed together; late fusion, where modalities are treated independently and their features or softmax classification scores are only combined in the very last layers; and intermediate feature fusion, where feature sharing is performed at middle layers of the network and hence the feature representations of different modalities are learnt jointly. 

A notable set of approaches in multimodal fusion rely on utilization of self-attention \cite{vaswani}. Recall that self-attention is formulated via calculating the dot-product similarity in the latent space, where queries $\mathbf{q}$,  keys $\mathbf{k}$, and values $\mathbf{v}$ are learnt from input feature representation via learnable projection matrices $\mathbf{W}_q, \mathbf{W}_k, \mathbf{W}_v$, and an attented representation is calculated based on them: 

\begin{equation}\label{eq:tau-att}
 \mathbf{A}_n = softmax\left(\frac{\mathbf{q}\mathbf{k}^T}{\sqrt{d}}\right)\mathbf{v},
  \end{equation} 
where $d$ is the dimensionality of a latent space. Considering the task of fusion of two modalities $a$ and $b$, self-attention can be utilized as a fusion approach by calculating queries from modality $a$ and keys and values from modality $b$. This results in representation learnt from modality $a$ attending to corresponding modality $b$ and further applying the obtained attention matrix to the representation learnt from modality $b$. 

This idea has been extensively utilized for solving a plethora of tasks involving multimodal fusion. In the context of emotion/affect recognition, notable works include \cite{mult, icasspav, interspeech}. In \cite{mult} the authors propose a multimodal transformer-based architecture for unaligned multimodal language sequences and consider fusion of three modalities, namely, audio, vision, and text. Data from each modality is first projected via a 1D convolutional layer to the desired dimensionality, and further, a set of transformer blocks is applied. Specifically, two transformer modules are utilized in each modality branch, with each of the two modules being responsible for fusion of one of the other modalities with the modality of the given branch following the above-specified approach. These representations are subsequently concatenated and another transformer module is applied in each modality branch on joint representations. The processed representations from each of three branches are subsequently concatenated for final classification.

Another relevant work is \cite{icasspav} where audio and visual modalities are considered for the task of emotion recognition. There, each modality is first preprocessed with a separate transformer block and representations learnt from each modality are fused with a transformer in a manner similar to Eq. (1). Compared to previous work, only one modality is fused into the other one, rather than performing fusion in a pair-wise manner in separate branches. 

Another work described in \cite{interspeech} considers emotion recognition from speech and text modalities. Similarly, they first perform modality-specific feature learning by means of convolutional blocks, and further employ two cross-modal attention blocks with one fusion audio into text, and the other one performing fusion into opposite direction. The statistics is pooled from each branch and concatenated for prediction.

As can be noticed, all the aforementioned methods are rather similar in their fusion strategies in that the transformer fusion is the building block of each of them, and the differences between the architectures are rather nominal and dataset-specific. At the same time, it can be noted that the models focus on building multimodal fusion methods rather than end-to-end emotion recognition systems, and often employ features that require separate estimation, especially for the vision modality. For example, \cite{mult} and \cite{icasspav} rely on facial action units as features, and \cite{mult} and \cite{interspeech} utilize language modality which requires separate annotation or estimation in practical application.

It should also be noticed that all three mentioned methods perform fusion in early or intermediate stages in the pipeline, forcing joint representations to be learnt. While benefiting from the joint feature learning, such fusion can become a curse if the learnt fused representations are too co-dependent and one of the modalities is noisy, incomplete, or simply non-existing during inference. Indeed, common practice has been to only evaluate the performance of the models under the ideal scenario of both modalities being present and complete at all times, while real-world applications do not necessarily reflect such scenarios.

\section{Proposed methods}
\begin{figure*}[h]
\includegraphics[width=0.95\textwidth]{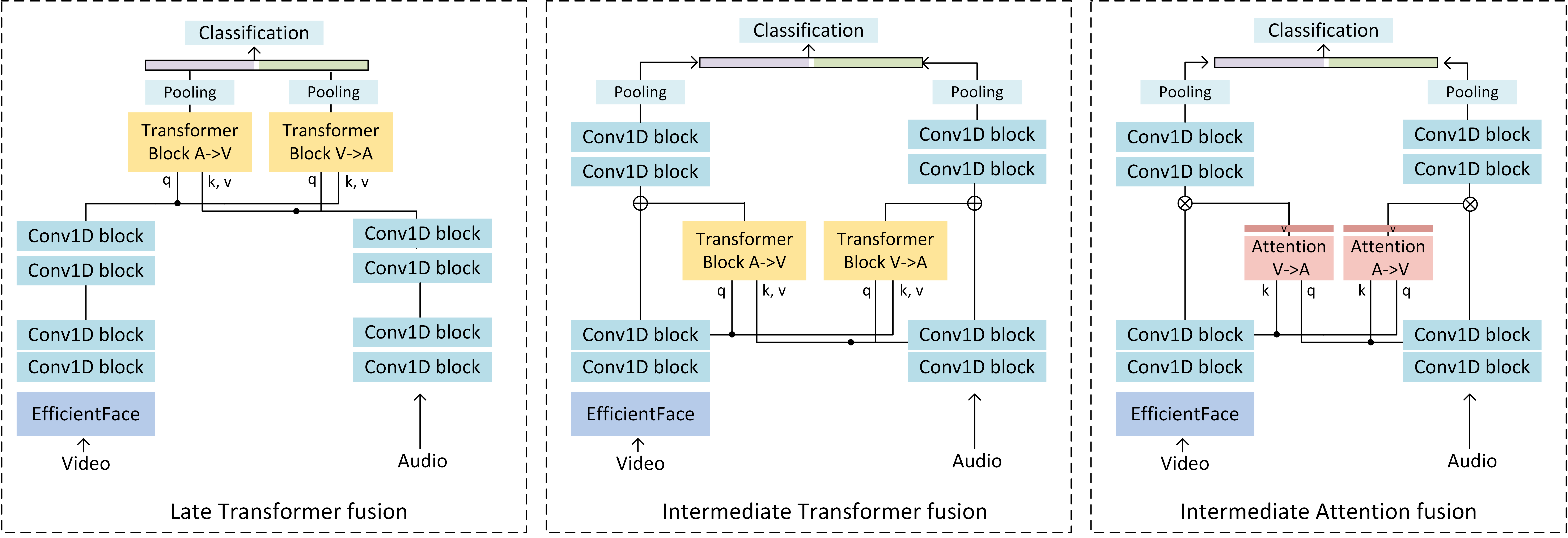}
\caption{Multimodal data fusion approaches.}
\end{figure*}

Here we describe the overall architecture of the proposed audiovisual emotion recognition model as well as three self-attention based modality fusion methods. Further, we propose an approach for accounting for missing data in one modality during inference in a form of modality dropout. On a general level, the model consists of two branches responsible for learning audio and visual features, respectively, and fusion modules placed either in the end or in the middle of the two branches depending on the feature fusion type, as shown in Figure 1. In both audio and visual branches, the 1D Convolutional blocks that are applied in a temporal dimension are primarily used.

\subsection{Feature extraction}

\subsubsection{Vision branch}

The vision branch consists of two parts, with the first part being the visual feature extraction from individual video frames, followed by learning of joint representation for the whole video sequence. To achieve an end-to-end trainable model capable of learning from raw video, we employ feature extraction as part of our pipeline and optimize it jointly with the multimodal fusion module, unlike the vast majority of existing works that separate feature extraction from multimodal fusion and mostly utilize pre-extracted features, such as facial landmark locations, facial action units, or head pose information \cite{mult, icasspav}. We choose one of the recently proposed facial expression recognition architectures, namely, EfficientFace \cite{efficientface} and incorporate it for feature extraction from individual frames prior to introducing them to subsequent 1D convolutional blocks. Specifically, the 1D convolutional blocks are added after average-pooled output of the last convolutional block of EfficientFace.

Considering an input video sequence of $k$ frames, each of the $k$ frames is processed independently by a 2D feature extractor, resulting in a single vector descriptor for each frame. These representations are further concatenated and processed in a temporal dimension with the temporal convolution blocks described further. We choose to follow this approach, as opposed to directly employing 3D-convolutions as commonly done in video tasks, as it provides a number of advantages in the given task, with the first one being the lower computational overhead brought by 2D convolutional layers compared to 3D convolutions. It can also be argued that temporal relations are of less importance in emotion recognition task, hence 1D convolutional operations applied in temporal dimension are sufficient to capture this information. Another major benefit of following the proposed approach is the ability to employ 2D feature extractor pre-trained on larger image-based emotion recognition datasets, as such datasets are significantly less common for videos that are necessary for pre-training 3D-convolutional models. 

Although we are primarily interested in building an end-to-end pipeline that can learn from raw data, the first part, i.e., visual feature extraction can be decoupled from the model and any other features can be used instead as input to the second part of the vision branch. That is, in the second part of the architecture, we assume that certain feature representation $\mathbf{X}_v^{N \times d}$ has been extracted from input visual data, where $N$ denotes the temporal dimension, and $d$ denotes the feature dimension. Here $\mathbf{X}_v$ can be represented by any feature types, either deep features extracted using a pre-trained model, or other features commonly used for emotion recognition, such as facial action units or landmarks. We further apply a sequence of four convolutional blocks for learning a temporal representation. Each convolutional block consists of an 1D Convoluitonal layer with a $3 \times 3$ kernel, Batch Normalization, and a ReLU activation. Further details can be seen in Table 1 that provides full details of vision branch, where $k$ denotes the kernel size, $d$ denotes the number of filters in a convolutional layer, and $s$ denotes the stride. The convolutional blocks are grouped into two stages for multimodal fusion described further.

\begin{table}[]
\centering
\begin{tabular}{|l|c|}
\multicolumn{2}{c}{Architecture of the visual branch} \\
\hline
     & EfficientFace module                              \\
\hline
\multirow{3}{*}{Stage1} & Reshape                                           \\
                        & Conv1D {[}k=3, d=64, s=1{]} + 
                        BN1D + ReLU  \\
                        & Conv1D {[}k=3, d=64, s=1{]} +
                        BN1D + ReLU  \\
                        \hline
\multirow{2}{*}{Stage2} & Conv1D {[}k=3, d=128, s=1{]} +
                         BN1D + ReLU \\
                        & Conv1D {[}k=3, d=128, s=1{]} + 
                         BN1D + ReLU \\ \hline
Predict                & Global Average Pooling + Linear        \\ \hline          
\multicolumn{2}{c}{ } \\
\multicolumn{2}{c}{Architecture of the audio branch} \\ \hline
\multirow{2}{*}{Stage1} & Conv1D {[}k=3, d=64{]} + BN1D + ReLU + MaxPool1d [2x1]                                   \\ 
                        & Conv1D {[}k=3, d=128{]} + BN1D + ReLU + MaxPool1d [2x1]    \\ \hline                                
\multirow{2}{*}{Stage2}                  & Conv1D {[}k=3, d=256{]} + BN1D + ReLU + MaxPool1d [k=2]                                    \\
                        & Conv1D {[}k=3, d=128{]} + BN1D + ReLU + MPool1D [k=2]                                   \\
                        \hline
Predict                 & Global Average Pooling + Linear                  \\ \hline
\end{tabular}
\caption{Architecture of the visual and audio modules}
\end{table}

\subsubsection{Audio branch}
Similarly to the vision branch, the audio branch operates on a feature representation, whether pre-computed or optimized jointly, and applies four blocks of 1D convolutional layers. Each block consists of a Convolutional layer, Batch Normalization, ReLU activation, and MaxPooling, with the specifications defined in Table 1.
For audio, we primarily use mel-frequency cepstral coefficients as features. We observed no benefit in using other feature representation types, such as chroma features or spectrograms. 

\subsection{Modality fusion approaches}

In this section, we describe the considered fusion approaches. We will first describe the late transformer fusion approach that is similar to previous works described in the literature, and then describe the two proposed intermediate fusion approaches. 

\subsubsection{Late transformer fusion}

In this setup, features learnt from two branches are fused with a transformer block. Specifically, we employ two transformers at the outputs of each branch, where fusion of one modality is performed into the other one. The outputs of these transformer blocks are further concatenated and passed to the final prediction layer. Formally, this can be defined as follows.

Let $\bm{\Phi}_a$ and $\bm{\Phi}_v$ be the feature representations of audio and vision modalities after the second feature extraction stage, i.e., after the fourth convolutional block. A transformer block is added in each branch, taking representations of two modalities as inputs. Considering the audio branch as an example, the transformer block takes the vision branch representation $\bm{\Phi}_v$ as input and projects it to obtain keys and values, while queries are computed from the audio branch features $\bm{\Phi}_a$. That is, self-attention is calculated as
\begin{equation}
A = softmax\left(\frac{ \bm{\Phi}_a\mathbf{W}_q\mathbf{W}_k^T \bm{\Phi}_v^T}{\sqrt{d}}\right)\bm{\Phi}_v\mathbf{W}_v,
\end{equation}
followed by standard transformer block processing \cite{vaswani}. The specific architecture of the transformer block is outlined in Figure 2.

\begin{figure}[h]
\centering
\includegraphics[width=0.3\textwidth]{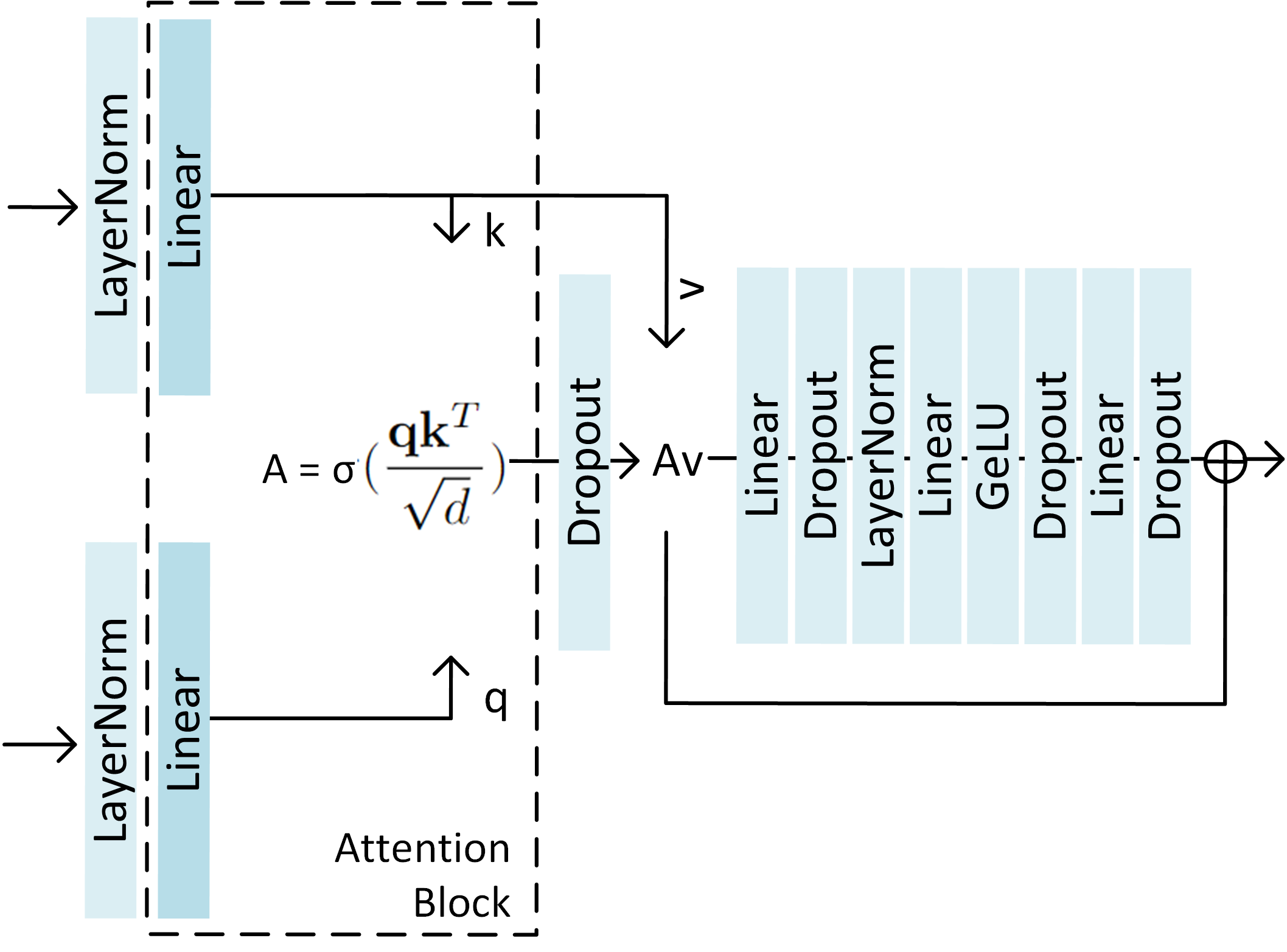}
\caption{Structure of the Transformer block.}
\end{figure}

The outputs of the two transformer blocks are concatenated and passed to the final layer for prediciton.

\subsubsection{Intermediate transformer fusion}

We propose the utilization of similar to the above-described transformer blocks for fusion at intermediate feature layers. Specifically, fusion is performed with a transformer block in each branch after the first stage of feature extraction, i.e., after two convolutional layers. Similar architecture to the one described in Figure 1 is used, and the fused feature representation is added to the corresponding branch.

Since data from complementary modality is introduced already at the earlier stage of the architecture, this allows to learn the features that are jointly meaningful for the task at hand between modalities during later convolutional layers.

\subsubsection{Intermediate attention-based fusion}

We further propose a fusion approach that is based merely on dot-product similarity that constitutes the attention in the transformer block. Formally, this is defined as follows. Given the two feature representations of different modalities $\bm{\Phi}_a$ and $\bm{\Phi}_b$, we compute queries and keys with learnt weights, similarly to conventional attention. The scaled dot-product similarity is subsequently calculated as
\begin{equation}
A = softmax\left(\frac{ \bm{\Phi}_a\mathbf{W}_q\mathbf{W}_k^T \bm{\Phi}_v^T}{\sqrt{d}}\right).
\end{equation}
Softmax activation promotes competition in the attention matrix, hence highlighting more important attributes/timestamps of each modality, and as a result providing the importance score of each key with respect to each query, i.e., each representation of modality $a$ with respect to modality $b$. This allows to calculate the relative importance of each attribute of modality $a$ by aggregating the scores corresponding to all the attributes of modality $b$ for each attribute of modality $a$. As a result, we obtain an attention vector that can be used to highlight more relevant attributes of the modality $a$. Considering the dot-product attention between features of audio and vision modalities shown in Equation 3, attention vector of vision modality is given by $\mathbf{v}_v = \sum_{i=N_v}A[:,i].$

Note that such a fusion approach does not directly fuse features of the two modalities. Instead, it identifies the attributes within each modality that are most relevant based to their similarity scores with data of the other modality. As a result, features that agree between the two modalities contribute the most to the final prediction, hence guiding the model towards learning modality-agnostic features or features with high level of agreement between the modalities. Such approach enables sharing of information between modalities, while not enforcing strong co-dependency of the learnt features in different branches as only attention scores are used for fusion. 

\subsection{Modality dropout}
The vast majority of the multimodal learning methods described to date assume the presence of both modalities at all times during inference. Nevertheless, oftentimes in real-world applications data of one or more modalities might not be reliable or may be missing at times. In such scenarios, conventional multi-modal approaches tend to fail. Here, we aim to account for the potential cases of missing data and propose the modality dropout as a way of mitigating it. As will be shown further, utilization of this approach leads to improved performance also in situations where both modalities are present. 

We propose the modality dropout, which randomly masks out or attenuates data of one of the modalities during training. Specifically, we consider three variants. In the first variant, during training, data of one modality in each sample is randomly selected and replaced with zeros, while the representation of the other modality for a given sample is kept intact. This approach imitates missing data and can also act as a regularizer similarly to Dropout layer utilized in neural networks. Note that in the case of the third fusion approach and absence of bias terms, this results in zero dot similarity matrix in the attention block, which after softmax and summation leads to constant attention vector, hence no information transfer from the zeroed modality. 

In the second variant, for each pair of data samples we generate a random scaling factor $\alpha$ in the range [0,1] \cite{shen2017continuous} and multiply one of the modalities by $\alpha$, while the other with $1 - \alpha$. The goal of this approach is to attenuate signals from different modalities at different training steps, and hence prevent the model from learning from strictly one modality. We further refer to this approach as `soft' modality dropout. The third variant is aimed at the problem of noisy data, where the input signal of one modality is corrupted. Here, the masking is performed similarly to the first variant, except rather than zero-masking, the data is randomly generated from a normal distribution with zero mean and unit variance in one of the modalities for each sample.

\section{Experiments}

In this section we describe the experimental protocol and the data used for assessing the performance of the proposed approaches. We report the results of the proposed model with three fusion variants, as well as recent multimodal emotion recognition methods, namely MULT \cite{mult} and multimodal transformer \cite{icasspav}. Note that both methods report results on datasets consisting of pre-extracted features. In addition, \cite{mult} considers three modalities, and \cite{icasspav} does not provide details on specific hyperparameters of the architecture, making direct comparison infeasible. We therefore adopt our feature extraction and compare with competing works only in terms of the fusion approaches described in these works. Specifically, to compare with \cite{icasspav} we employ a transformer block on top of our two convolutional branches that performs fusion either from audio to video, or in the opposite direction, and replace linear layers in the transformer block with 1D-convolutional ones. Regarding MULT \cite{mult}, we want to compare with purely audiovisual model, so we remove the transformer blocks responsible for fusion from/to the language modality. This yields the architecture that is similar to our late transformer fusion, with additional single modality transformer blocks added in each branch. Other hyperparameters, such as latent space dimensionality, are kept identical between the methods for fair comparison. Similarly to the comparison with \cite{mult}, we add our feature extraction blocks to the model. Unless otherwise specified, we use a single transformer block with single head everywhere to achieve a lightweight model. Naturally, better performance can be expected from adding additional blocks and parameters to the models. We used \cite{timm} for transformer block implementation.

Subsequently, we perform experiments with modality dropout in two settings. In the first one, we target the problem of missing data from one modality and apply both soft and hard modality dropout during training. That is, in this setting, in each batch the data consists of the pairs of full data, pairs without audio modality, pairs without video modality, and pairs multiplied with random coefficients as described above. We report the performance in the presence of both modalities (denoted by `AV'), as well as in the full absence of one modality (denoted by `A' or `V' for presence of only audio and video modalities, respectively). We additionally report the average metric over the three modality settings (denoted by `M') to simplify the comparison between methods. In the second setting, we consider robustness towards noise and apply the third variant of modality dropout during training, and replace one of the modalities with random noise during testing. 

\subsubsection{RAVDESS dataset}
We choose RAVDESS dataset \cite{ravdess} primarily due to availability of raw data in this dataset, as opposed to others. The dataset consists of video recordings of 24 people speaking with different emotions and poses a task of classification of emotional states into 7 classes: calm, happy, sad, angry, fearful, surprise, and disgust. 60 video sequences were recorded for each actor, and we crop or zero-pad them to 3.6 seconds, which is the average sequence length. For audio processing, we extract 10 Mel-frequency cepstral coefficients for further processing. 
For visual data, we select 15 uniformly distributed frames from 3.6 second video, and crop the faces of actors using a face detection algorithm \cite{gradilla2020multi}. Images are resized into 224x224 pixels. We train the model on raw 15-frame videos. We transfer the weights of EfficientFace pre-trained on AffectNet dataset \cite{mollahosseini2017affectnet}. 
We split the data into training, validation and test sets ensuring that the identities of actors are not repeated across sets. Specifically, we used four actors for testing, four for validation, and 16 for training, and report the result averaged over five folds. The videos are scaled into [0,1] scale, and random horizontal flip and random rotation are used for data augmentation. All the models are trained for 100 epochs with SGD, learning rate of 0.04, momentum of 0.9, weight decay of 1e-3, and reduction of learning rate on plateau of 10 epochs.

\subsubsection{CMU-MOSEI dataset}
We additionally conduct experiments on CMU-MOSEI dataset. The dataset consists of 23,454 movie review video clips taken from YouTube and labeled by human annotators with a sentiment score in the range [-3..3]. Note that we only consider audio and visual modalities in our experiments. Since the dataset provides pre-extracted features rather than raw data (specifically, 35 facial action units are provided for vision modality and audio data is represented by mfccs, pitch tracking, glottal source and peak slope parameters resulting in 74 features), we omit the EfficientFace feature extraction in the vision branch and training is performed starting from convolutional blocks directly. We rely on the implementation of \cite{mult} for the experimental protocol on CMU-MOSEI dataset and adopt the training hyperparameters described in therein.

\begin{table}[h]
\scriptsize

\setlength{\tabcolsep}{3pt}

\begin{tabular}{|p{12pt}|p{12pt}p{13pt}p{13pt}p{14pt}|p{12pt}p{13pt}p{13pt}p{14pt}|p{12pt}p{13pt}p{13pt}p{14pt}|}
\hline
\multicolumn{1}{|l|}{}         & \multicolumn{4}{c|}{RAVDESS. ACC}                                                      & \multicolumn{4}{c|}{MOSEI. ACC}                                                         & \multicolumn{4}{c|}{MOSEI. MAE}                                                        \\ \hline
\multicolumn{1}{|l|}{}   & AV             & A              & V              & {M}              & AV             & A              & V               & {M}              & AV             & A              & V              & {M}              \\
\multicolumn{1}{|l|}{LT1} &      \textbf{79.33}
          &        19.83
        &        36.41
        & \textbf{45.19}               & 63.89          & 48.70          & 62.85           & {58.48}          & 0.806          & 0.840          & 1.063          & {0.903}          \\
\multicolumn{1}{|l|}{LT4} & 76.42          & \textbf{27.92}          & \textbf{30.00}          & 44.78          & 66.56          & 62.63          & 53.16           & 60.78          & 0.806          & 0.839          & 0.831          & 0.825          \\
\multicolumn{1}{|l|}{IT1}    &76.41
         & 21.16
          & 18.33
         & 38.63      & \textbf{67.72} & 37.14          & 62.87           & 55.91         & \textbf{0.792} & 0.843          & 0.809          & 0.815 \\
\multicolumn{1}{|l|}{IT4}    &     78.50
           &       20.33
         &    17.33
            & 38.72               & 64.91          & 62.60          & 62.85           & 63.45        & 0.817          & 0.840          & 0.832          & 0.830         \\
\multicolumn{1}{|l|}{IA1}    &      76.00         &       18.58         &     22.83
           &
 39.13               & 64.94          & 62.08          & 62.86           & 63.29          & 0.802          & \textbf{0.837} & 0.806          & 0.815 \\
\multicolumn{1}{|l|}{IA4}    &     77.41
           &     20.66
           &      29.83
          & 42.63               & \textbf{67.72}          & \textbf{63.07}          & \textbf{65.77}           & \textbf{65.52}     & 0.794          & \textbf{0.837}          & \textbf{0.803}         & \textbf{0.811}      \\
\multicolumn{1}{|l|}{TAV}      & 77.75          & 24.25          & 13.33          & 38.44        & 64.94          & 62.08          & 62.86           & 62.18       & 0.814          & 0.841          & 1.093          & 0.916        \\
\multicolumn{1}{|l|}{TVA}      & 76.00          & 15.16          & 42.67          & 44.61       & 66.48          & 37.15          & 56.96           & 53.53        & 0.809          & 0.852          & 0.838          & 0.833        \\
\multicolumn{1}{|l|}{MLT }    & 74.16          & 22.33          & 35.42          & 43.97       & 62.90          & 62.85 & 64.44  & 63.40 & 0.804          & 0.838          & 0.804 & 0.815 \\ \hline
\multicolumn{13}{c}{MODALITY DROPOUT}                                                                                                                                                                                                                                                                     \\ \hline
\multicolumn{1}{|l|}{LT1} & 79.08               &      59.16          &   72.66             & {70.30}               & 67.11          & 63.62          & 0.629           & {64.54}          & 0.802          & 0.829          & 0.801          & {0.811}          \\
\multicolumn{1}{|l|}{LT4} & 79.25          & 53.00          & 70.92          & {67.72}          & 64.47          & 53.71          & 64.91           & {61.03}          & 0.814          & 0.837          & 0.819          & {0.824}          \\
\multicolumn{1}{|l|}{IT1}    &     77.33           & 48.41     &  73.75                & {66.50}               & 62.80          & 62.85          & 63.09           & {62.91}          & 0.804          & 0.831          & 0.803          & {0.813}          \\
\multicolumn{1}{|l|}{IT4}    & 78.91          & 44.33          & \textbf{74.92} & {66.05}          & 67.01          & 64.30          & 63.12           & {64.81}          & 0.796          & 0.826          & \textbf{0.797}          & {0.806}          \\
\multicolumn{1}{|l|}{IA1}    & \textbf{81.58} & 58.08          & 72.83          & {\textbf{70.83}} & \textbf{67.19} & \textbf{64.52} & \textbf{64.91} & {\textbf{65.54}}          & 0.795          & \textbf{0.816} & 0.798 & {\textbf{0.803}} \\
\multicolumn{1}{|l|}{IA4}    &      79.58          &     57.16           &    71.83            & {69.52}               & 63.48          & 62.74          & 63.18           & {63.13}          & 0.807          & 0.820          & 0.808          & {0.812}          \\
\multicolumn{1}{|l|}{TAV}      & 76.58          & 54.83          & 13.33          & {48.24}          & 65.32          & 63.84        & 62.85           & {64.01}          & 0.811          & 0.832          & 0.839          & {0.828}          \\
\multicolumn{1}{|l|}{TVA}      & 74.42          & 44.91          & 69.58          & {62.97}          & 67.61        & 63.98          & 60.95           & {64.18}          & \textbf{0.793}          & 0.819          & 0.798          & {0.803}          \\
\multicolumn{1}{|l|}{MLT}     & 78.50          & 53.58          & 70.66          & {67.58}          & 63.87          & 62.85          & 63.37           & {63.36}          & 0.806 & 0.836          & 0.835          & {0.826}          \\ \hline 
\multicolumn{13}{c}{MODALITY DROPOUT with NOISE}                                         \\ \hline
\multicolumn{1}{|l|}{LT1} & 77.08          & 53.16          & 68.50          & {66.246}         & 65.57          & 64.03          & 64.94           & {64.94}          & 0.809          & 0.826          & 0.806          & {0.813}          \\
\multicolumn{1}{|l|}{LT4} & 80.33          & 54.33          & 73.00          & {69.22}          & 64.08          & 63.31          & 62.85           & {62.85}          & 0.813          & 0.827          & 0.813          & {0.818}          \\
\multicolumn{1}{|l|}{IT1}    & 76.75          & 53.75          & 71.58          & {67.36}          & \textbf{68.16} & \textbf{65.98} & 63.53           & {63.53}          & \textbf{0.799} & 0.821          & 0.804          & {\textbf{0.808}} \\
\multicolumn{1}{|l|}{IT4}    & 76.08          & 54.50          & 71.00          & {67.19}          & 67.83          & 63.56          & 64.22           & {64.22}          & 0.801          & 0.826          & \textbf{0.802} & {0.809}          \\
\multicolumn{1}{|l|}{IA1}    & 78.25          & \textbf{58.25} & \textbf{71.66} & {\textbf{69.38}} & 62.76          & 63.89          & 63.18           & {63.27}          & 0.804          & \textbf{0.819} & 0.805          & {0.809}          \\
\multicolumn{1}{|l|}{IA4}    & \textbf{78.41} & 55.75          & 68.58          & {67.58}          & 63.51          & 64.08          & 62.54           & {63.37}          & 0.805          & 0.820          & 0.808          & {0.811}          \\
\multicolumn{1}{|l|}{TAV}      & 75.83          & 56.25          & 12.83          & {48.30}          & 66.81          & 65.68          & \textbf{65.60}   & {\textbf{66.03}} & 0.810          & 0.820           & 0.811          & {0.813}          \\
\multicolumn{1}{|l|}{TVA}      & 73.66          & 41.25          & 71.41          & {62.10}          & 66.23          & 63.18          & 64.58           & {64.66}          & 0.804          & 0.831          & 0.806          & {0.813}          \\
\multicolumn{1}{|l|}{MLT}     & 77.41          & 54.16          & 66.33          & {65.96}          & 64.52          & 62.74          & 63.51           & {63.59}          & 0.805          & 0.830          & 0.805          & {0.811}          \\ \hline
\end{tabular}
\caption{Performance of different fusion methods on RAVDESS and MOSEI.}
\end{table}
\begin{table}[]
\scriptsize
\centering
\begin{tabular}{|l|p{13pt}p{13pt}p{13pt}p{13pt}|p{13pt}p{13pt}p{13pt}p{13pt}|}
\hline
       & \multicolumn{4}{c|}{MOSEI. ACC}                                    & \multicolumn{4}{|c|}{MOSEI. MAE}                                    \\
       \hline
       & AV             & A              & V              & M              & AV             & A              & V              & M              \\
IT1    & 64.66          & 38.80          & 63.12          & 55.53          & 0.821          & 0.857          & \textbf{0.803} & 0.827          \\
IT4    & 65.90          & 37.23          & 62.85          & 55.32          & 0.805          & 0.845          & 1.932          & 1.194          \\
IA1    & 64.09          & \textbf{62.85} & \textbf{63.42} & \textbf{63.45} & 0.799          & \textbf{0.838} & 0.807          & \textbf{0.815} \\
IA4    & 64.74          & 37.28          & 61.28          & 54.43          & 0.803          & 0.842          & 0.808          & 0.818          \\
MLT   & \textbf{67.66} & 56.90          & 60.73          & 61.76          & \textbf{0.787} & \textbf{0.838} & 0.836          & 0.821          \\
\hline
\multicolumn{9}{c}{MODALITY DROPOUT}                                                                    \\
\hline
IT1  & 65.41          & 62.85          & 64.06          & 64.11          & 0.805          & 0.838          & 0.805          & 0.816          \\
IT4  & 66.57          & 64.78          & \textbf{65.02} & 65.45          & 0.792          & \textbf{0.812} & \textbf{0.795} & \textbf{0.800} \\
IA1  & \textbf{68.76} & \textbf{65.96} & 63.92          & \textbf{66.21} & \textbf{0.791} & 0.815          & 0.799          & 0.802          \\
IA4  & 66.18          & 64.67          & 64.22          & 65.02          & 0.794          & 0.815          & 0.801          & 0.803          \\
MLT   & 66.12          & 65.41          & 63.62          & 65.05          & 0.801          & 0.831          & 0.808          & 0.813          \\
\hline
\multicolumn{9}{c}{MODALITY DROPOUT with NOISE}                                                                                                     \\
\hline
IT1  & 64.72          & 54.07          & \textbf{66.34} & 61.71          & 0.798          & 0.839          & 0.797          & 0.812          \\
IT4  & 64.69          & 64.33          & 61.83          & 63.61          & 0.801          & 0.826          & 0.799          & 0.808          \\
IA1  & \textbf{67.25} & \textbf{64.96} & 64.74          & \textbf{65.65} & 0.794         & \textbf{0.813} & 0.799          & 0.802          \\
IA4  & 63.40          & 63.23          & 62.85          & 63.16          & 0.806 & 0.820          & 0.806          & 0.811          \\
MLT   & 66.18          & 64.19          & 64.39          & 64.92          & \textbf{0.790}          & \textbf{0.813} & \textbf{0.791} & \textbf{0.798} \\ \hline 
\end{tabular}
\caption{Comparison with MULT \cite{mult}.}
\end{table}

\subsection{Results and Discussion}

Table II shows the results of the proposed approaches on the RAVDESS and MOSEI datasets. Here, `LT1' and `LT4' denote late transformer fusion with one and four heads, respectively, and similarly `IT' denotes intermediate transformer fusion, `IA' denotes intermediate attention fusion, `TAV' and `TVA' refers to the fusion approaches described in \cite{icasspav}, and `MULT' refers to \cite{mult}. We report categorical accuracy on RAVDESS dataset, and binary accuracy (positive vs negative sentiment) on MOSEI dataset, as well as Mean Average Error between the true and predicted sentiment scores. 

As can be seen, in the setting without any type of dropout, late transformer fusion achieves the best result on RAVDESS dataset, while intermediate attention fusion achieves the best result on MOSEI dataset on both the accuracy and MAE metrics. Note that intermediate attention fusion is also the most lightweight fusion approach compared to any of the methods using full transformer blocks. At the same time, performance under the presence of only one modality is extremely poor on RAVDESS dataset. On MOSEI dataset the performance drop is not drastic in the majority of cases, likely due to the dataset consisting of already pre-extracted features, and hence guaranteeing presence of meaningful independent features in each modality even in the absence of the other one. 

Further, it can be seen that utilization of modality dropout improves the performance drastically under incomplete data of one modality. This is the case for all fusion methods, while intermediate attention fusion benefits from it the most.  Besides, the performance under the presence of both modalities is improved as well, with the best result on RAVDESS achieved by intermediate attention fusion. This is also the best result on this dataset among all methods and dropout settings. Similar conclusions can be made on MOSEI dataset; utilization of modality dropout improves the performance in both single-modality and two-modality case. 
Under the noisy setting, we still observe the intermediate attention fusion performing best on the average metric on RAVDESS. 

To provide better comparison with state-of-the-art, we additionally compare with full MULT model (omitting the language modality), following the implementations provided by \cite{mult} and using their convolutional layers, transformer block implementations and other hyperparameters. Since in their implementation several dense layers are added after the fusion and prior to the output layer, we add similar dense layers to our model for fair comparison. The results are provided in Table III. As can be seen, while MULT outperforms the proposed intermediate fusion approaches in the vanilla setting with both modalities, intermediate fusion handles missing modalities better, and especially under the presence of modality dropouts. The best overall performance is achieved by intermediate attention fusion with the first modality dropout variant.  

As can be seen, in the majority of the cases the best performance is achieved by the proposed intermediate attention fusion combined with one of the proposed dropout approaches. As in this approach no hard feature sharing is performed, the learnt feature representations are less likely to be co-dependent and therefore can be disentangled more easily, hence leading to better robustness of the model in incomplete data settings. This, in turn, leads to better generalization capabilities of the model overall, leading to improved performance also under the setting of both modalities.

\section{Conclusion}
We proposed a model for audiovisual emotion recognition that learns end-to-end and an attention-based fusion method. We evaluated the robustness of different modality fusion approaches under the absence of, or noise present in, one of the modalities and proposed an approach to improve the model's robustness. Importantly, the proposed approach also improves the performance under the (ideal) standard setting where both modalities are present.






\bibliographystyle{IEEEtran}
\bibliography{references}

\begin{thebibliography}{10}
\providecommand{\url}[1]{#1}
\csname url@samestyle\endcsname
\providecommand{\newblock}{\relax}
\providecommand{\bibinfo}[2]{#2}
\providecommand{\BIBentrySTDinterwordspacing}{\spaceskip=0pt\relax}
\providecommand{\BIBentryALTinterwordstretchfactor}{4}
\providecommand{\BIBentryALTinterwordspacing}{\spaceskip=\fontdimen2\font plus
\BIBentryALTinterwordstretchfactor\fontdimen3\font minus
  \fontdimen4\font\relax}
\providecommand{\BIBforeignlanguage}[2]{{%
\expandafter\ifx\csname l@#1\endcsname\relax
\typeout{** WARNING: IEEEtran.bst: No hyphenation pattern has been}%
\typeout{** loaded for the language `#1'. Using the pattern for}%
\typeout{** the default language instead.}%
\else
\language=\csname l@#1\endcsname
\fi
#2}}
\providecommand{\BIBdecl}{\relax}
\BIBdecl

\bibitem{liu2017facial}
Z.~Liu, M.~Wu, W.~Cao, L.~Chen, J.~Xu, R.~Zhang, M.~Zhou, and J.~Mao, ``A
  facial expression emotion recognition based human-robot interaction system,''
  \emph{IEEE/CAA Journal of Automatica Sinica}, vol.~4, no.~4, pp. 668--676,
  2017.

\bibitem{chen2019automatic}
J.~Chen, Y.~Lv, R.~Xu, and C.~Xu, ``Automatic social signal analysis: Facial
  expression recognition using difference convolution neural network,''
  \emph{Journal of Parallel and Distributed Computing}, vol. 131, pp. 97--102,
  2019.

\bibitem{dzedzickis2020human}
A.~Dzedzickis, A.~Kaklauskas, and V.~Bucinskas, ``Human emotion recognition:
  Review of sensors and methods,'' \emph{Sensors}, vol.~20, no.~3, p. 592,
  2020.

\bibitem{mollahosseini2017affectnet}
A.~Mollahosseini, B.~Hasani, and M.~H. Mahoor, ``Affectnet: A database for
  facial expression, valence, and arousal computing in the wild,'' \emph{IEEE
  Transactions on Affective Computing}, vol.~10, no.~1, pp. 18--31, 2017.

\bibitem{soleymani2017survey}
M.~Soleymani, D.~Garcia, B.~Jou, B.~Schuller, S.-F. Chang, and M.~Pantic, ``A
  survey of multimodal sentiment analysis,'' \emph{Image and Vision Computing},
  vol.~65, pp. 3--14, 2017.

\bibitem{khalil2019speech}
R.~A. Khalil, E.~Jones, M.~I. Babar, T.~Jan, M.~H. Zafar, and T.~Alhussain,
  ``Speech emotion recognition using deep learning techniques: A review,''
  \emph{IEEE Access}, vol.~7, pp. 117\,327--117\,345, 2019.

\bibitem{efficientface}
Z.~Zhao, Q.~Liu, and F.~Zhou, ``Robust lightweight facial expression
  recognition network with label distribution training,'' in \emph{Proceedings
  of the AAAI Conference on Artificial Intelligence}, vol.~35, no.~4, 2021, pp.
  3510--3519.

\bibitem{mmtm}
H.~R.~V. Joze, A.~Shaban, M.~L. Iuzzolino, and K.~Koishida, ``Mmtm: Multimodal
  transfer module for cnn fusion,'' in \emph{Proceedings of the IEEE/CVF
  Conference on Computer Vision and Pattern Recognition}, 2020, pp.
  13\,289--13\,299.

\bibitem{chumachenko2021speed}
K.~Chumachenko, J.~Raitoharju, A.~Iosifidis, and M.~Gabbouj, ``Speed-up and
  multi-view extensions to subclass discriminant analysis,'' \emph{Pattern
  Recognition}, vol. 111, p. 107660, 2021.

\bibitem{wang2020deep}
Y.~Wang, W.~Huang, F.~Sun, T.~Xu, Y.~Rong, and J.~Huang, ``Deep multimodal
  fusion by channel exchanging,'' \emph{Advances in Neural Information
  Processing Systems}, vol.~33, 2020.

\bibitem{xiao2020audiovisual}
F.~Xiao, Y.~J. Lee, K.~Grauman, J.~Malik, and C.~Feichtenhofer, ``Audiovisual
  slowfast networks for video recognition,'' \emph{arXiv preprint
  arXiv:2001.08740}, 2020.

\bibitem{xiao2020multimodal}
Y.~Xiao, F.~Codevilla, A.~Gurram, O.~Urfalioglu, and A.~M. L{\'o}pez,
  ``Multimodal end-to-end autonomous driving,'' \emph{IEEE Transactions on
  Intelligent Transportation Systems}, 2020.

\bibitem{ignore}
F.~Laakom, K.~Chumachenko, J.~Raitoharju, A.~Iosifidis, and M.~Gabbouj,
  ``Learning to ignore: rethinking attention in cnns,'' \emph{arXiv preprint
  arXiv:2111.05684}, 2021.

\bibitem{mult}
Y.-H.~H. Tsai, S.~Bai, P.~P. Liang, J.~Z. Kolter, L.-P. Morency, and
  R.~Salakhutdinov, ``Multimodal transformer for unaligned multimodal language
  sequences,'' in \emph{Proceedings of the conference. Association for
  Computational Linguistics. Meeting}, vol. 2019.\hskip 1em plus 0.5em minus
  0.4em\relax NIH Public Access, 2019, p. 6558.

\bibitem{interspeech}
D.~Krishna and A.~Patil, ``Multimodal emotion recognition using cross-modal
  attention and 1d convolutional neural networks.'' in \emph{Interspeech},
  2020, pp. 4243--4247.

\bibitem{icasspav}
J.~Huang, J.~Tao, B.~Liu, Z.~Lian, and M.~Niu, ``Multimodal transformer fusion
  for continuous emotion recognition,'' in \emph{ICASSP 2020-2020 IEEE
  International Conference on Acoustics, Speech and Signal Processing
  (ICASSP)}.\hskip 1em plus 0.5em minus 0.4em\relax IEEE, 2020, pp. 3507--3511.

\bibitem{shu2018review}
L.~Shu, J.~Xie, M.~Yang, Z.~Li, Z.~Li, D.~Liao, X.~Xu, and X.~Yang, ``A review
  of emotion recognition using physiological signals,'' \emph{Sensors},
  vol.~18, no.~7, p. 2074, 2018.

\bibitem{vaswani}
A.~Vaswani, N.~Shazeer, N.~Parmar, J.~Uszkoreit, L.~Jones, A.~N. Gomez,
  {\L}.~Kaiser, and I.~Polosukhin, ``Attention is all you need,'' in
  \emph{Advances in neural information processing systems}, 2017, pp.
  5998--6008.

\bibitem{shen2017continuous}
X.~Shen, X.~Tian, T.~Liu, F.~Xu, and D.~Tao, ``Continuous dropout,'' \emph{IEEE
  transactions on neural networks and learning systems}, vol.~29, no.~9, pp.
  3926--3937, 2017.

\bibitem{timm}
R.~Wightman, ``Pytorch image models,''
  \url{https://github.com/rwightman/pytorch-image-models}, 2019.

\bibitem{ravdess}
S.~R. Livingstone and F.~A. Russo, ``The ryerson audio-visual database of
  emotional speech and song (ravdess): A dynamic, multimodal set of facial and
  vocal expressions in north american english,'' \emph{PloS one}, vol.~13,
  no.~5, p. e0196391, 2018.

\bibitem{gradilla2020multi}
R.~Gradilla, ``Multi-task cascaded convolutional networks (mtcnn) for face
  detection and facial landmark alignment,'' \emph{link]. Acessado em},
  vol.~13, 2020.

\end{thebibliography}
%



\end{document}